\documentclass[sigconf,balance=false]{acmart}
\usepackage{subcaption}
\usepackage{algorithm}
\usepackage{algpseudocode}
\usepackage{multirow}
\usepackage{colortbl}
\definecolor{MainAccFirstColor}{HTML}{FEE3D6}
\definecolor{MainAccSecondColor}{HTML}{DAE3F0}
\definecolor{MainAccThirdColor}{HTML}{D8ECEF}
\newcommand{\MainAccFirst}[1]{\cellcolor{MainAccFirstColor}#1}
\newcommand{\MainAccSecond}[1]{\cellcolor{MainAccSecondColor}#1}
\newcommand{\MainAccThird}[1]{\cellcolor{MainAccThirdColor}#1}

\newcommand{\ResultTableStyle}{\small\setlength{\tabcolsep}{0.85pt}}

\AtBeginDocument{%
  }

\setcopyright{none}
\renewcommand\footnotetextcopyrightpermission[1]{}
\acmConference[WWW '27]{The ACM Web Conference 2027}{May 10--14, 2027}{Dublin, Ireland}
\acmYear{2027}
\acmDOI{}
\acmISBN{}

\begin{document}

\title{HeteGCD: Statistics-Aware Dual-Prototype Learning for Generalized Category Discovery on Heterophilous Graphs}

\settopmatter{authorsperrow=1}

\author{%
  Xuanzhi Xi, Tongze Zhang, Zhili Qin, Yulin Tao,\\
  Hongliang Wang, Zhong Zhang, Qinli Yang, Junming Shao
}

\affiliation[obeypunctuation=true]{%
  \institution{University of Electronic Science and Technology of China},
  \country{China}
}

\email{xuanzhixi@std.uestc.edu.cn}

\renewcommand{\shortauthors}{Xuanzhi Xi et al.}

\begin{abstract}
\begingroup
Heterophilous graphs are prevalent in real-world applications,\linebreak where connected nodes often exhibit dissimilar attributes or belong to different classes, posing significant challenges to conventional graph learning methods based on homophily.
This limitation becomes particularly pronounced in generalized category discovery (GCD), which requires models to recognize known classes and discover novel ones using labels from known classes only.
However, existing graph GCD methods typically rely on aforementioned homophily-based inductive biases, which are ill-suited to heterophilous graphs.
Meanwhile, although current heterophilous graph learning methods exploit diverse neighborhood signals, their selection and fusion are often guided by supervision across all classes, including novel classes whose labels are unavailable in GCD.
These limitations raise a key question: how can heterogeneous neighborhood information be effectively leveraged for category discovery without any novel-class supervision?
To answer this question, in this study we propose HeteGCD, a new statistics-aware dual-prototype learning perspective for generalized category discovery on heterophilous graphs.
The key idea is to examine both statistics of local attribute differences and neighborhood class distribution to extract discriminative information in feature and probability spaces simultaneously.
Specifically, a gated dual-stream encoder integrates both node attributes and local attribute-difference statistics to learn informative node representations.
Afterwards, a dual-prototype classifier combines feature-based similarity with distribution-based consistency to support category discovery.
Experiments on five heterophilous graphs show that HeteGCD achieves competitive performance against 24 baseline combinations.
\par
\endgroup
\end{abstract}


\maketitle

\section{Introduction}
\label{sec:introduction}

Graphs are widely used to represent relationships in real-world applications, ranging from social networks to online shopping platforms. Many graph learning methods build upon the homophily assumption that connected nodes tend to belong to the same class~\cite{zhu2020h2gcn}. However, many Web graphs exhibit heterophily, with frequent connections between nodes of different classes~\cite{platonov2023critical}. In product co-purchase networks, for example, customers may purchase mobile phones and chargers together, creating links between complementary products from different categories. Moreover, available annotations may cover only a subset of classes. An online store, for instance, may have labeled examples of phones and laptops, while a newly introduced category, such as virtual reality (VR) headsets, has no labeled examples yet. These practical challenges motivate the study of generalized category discovery (GCD) on heterophilous graphs, a problem that remains underexplored. Given labels for only some nodes from known classes, the goal is to recognize known classes and discover novel ones among the remaining unlabeled nodes~\cite{vaze2022gcd}.

\begin{figure}[t]
  \centering
  \includegraphics[width=\linewidth]{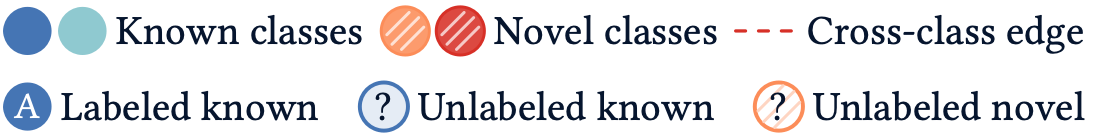}\par
  \begin{subfigure}[t]{.485\linewidth}
    \centering
    \includegraphics[width=\linewidth]{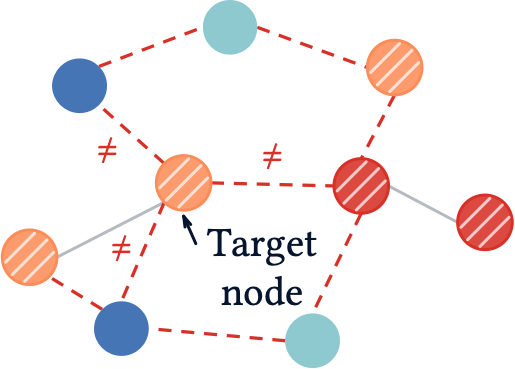}
    \caption{\textbf{Unreliable Homophily Assumption.} Nodes often connect to nodes of different classes, challenging graph GCD methods that rely on homophily.}
    \label{fig:intro_homophily}
  \end{subfigure}\hfill
  \begin{subfigure}[t]{.485\linewidth}
    \centering
    \includegraphics[width=\linewidth]{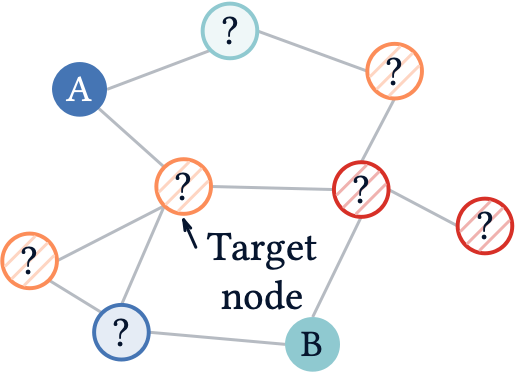}
    \caption{\textbf{Unavailable Novel-Class Supervision.} Missing novel-class feedback hinders signal selection and fusion for closed-set heterophilous methods.}
    \label{fig:intro_supervision}
  \end{subfigure}
  \caption{\textbf{GCD on Heterophilous Graphs.} Only a subset of known-class nodes is labeled, while unlabeled nodes include both known and novel classes.}
  \label{fig:intro_graph_gcd}
  \Description{Two panels show the same heterophilous graph with four known-class nodes, five novel-class nodes, two same-class edges, and nine cross-class edges. Panel a uses red dashed cross-class edges, gray solid same-class edges, and inequality signs beside the target node's three cross-class edges. Panel b uses gray edges. Two labeled known-class nodes are solid and carry labels A and B. The other known-class nodes have pale fills and question marks; novel-class nodes have white fills, colored outlines, light hatching, and question marks. The shared legend has two independently spaced rows of three items, with equally sized example nodes: known classes, novel classes, cross-class edge, labeled known, unlabeled known, and unlabeled novel. Colors indicate true classes only for illustration.}
\end{figure}

On the one hand, heterophily challenges the homophily assumption underlying existing graph GCD methods. Existing methods for graph-based category discovery and open-world node classification often rely on neighborhood agreement to select or propagate pseudo-labels~\cite{hoffmann2025pown,xi2026toponc}. However, as illustrated in Figure~\ref{fig:intro_graph_gcd}(a), most of the target node's neighbors belong to other classes, with red dashed edges highlighting these cross-class connections. Neighborhood agreement can therefore provide misleading evidence for category assignment, undermining the homophily-based inductive biases of these methods.

On the other hand, the lack of novel-class supervision challenges conventional heterophilous graph learning methods. These methods have achieved strong performance in closed-set node classification by combining node attributes with diverse neighborhood and structural signals~\cite{zhu2020h2gcn,bo2021fagcn,luan2022acm,li2022glognn,lim2021linkx}. Labels from all target classes guide these methods in learning which signals to use and how to combine them. However, in the GCD setting illustrated in Figure~\ref{fig:intro_graph_gcd}(b), only some known-class nodes are labeled, while the remaining unlabeled nodes, marked with question marks, belong to both known and novel classes. The model therefore lacks direct novel-class feedback to determine which signals help distinguish novel classes and how these signals should be combined.

When unreliable homophily priors meet missing novel-class supervision, a key question arises: \emph{how can heterogeneous neighborhood information be effectively used for category discovery without any novel-class supervision?}

To answer this question, we examine neighborhood information from a statistical perspective, with node attributes as the basis. We study two types of neighborhood statistics: \emph{local attribute-difference statistics}, which summarize a node's deviations from its neighbors' mean attributes, and \emph{neighborhood class distribution statistics}, which describe the proportions of classes in its neighborhood. Our empirical analyses provide evidence of category separability in these three signals, supporting their use for discovery. These observations suggest combining node attributes and local attribute-difference statistics in embedding space, while retaining neighborhood class distributions in probability space to guide category assignment.

Based on these observations, we propose \textbf{\mbox{HeteGCD}}, a statistics-aware dual-prototype learning framework. A gated dual-stream encoder integrates node attributes and local attribute-difference statistics to learn node representations. A dual-prototype classifier combines similarity to feature prototypes with consistency between neighborhood distributions and distribution prototypes. The neighborhood distributions and distribution prototypes are estimated from evolving soft assignments over known and novel classes, without requiring novel-class labels.

Our main contributions are as follows.
\begin{itemize}
  \item \textbf{Heterophilous graph GCD.} To the best of our knowledge, this is the first systematic study of GCD on heterophilous graphs, examining the joint challenges of unreliable homophily priors and unavailable novel-class supervision.

  \item \textbf{Statistics-Aware Dual-Prototype Learning.} We propose HeteGCD, which combines a gated dual-stream encoder with a dual-prototype classifier to capture category information from node attributes and neighborhood statistics in feature and probability spaces.

  \item \textbf{Competitive performance.} \mbox{HeteGCD} achieves the best average ranks in All and New accuracy across five heterophilous graphs when compared with 24 baseline combinations.
\end{itemize}

\section{Related Work}
\label{sec:related-work}

\subsection{Generalized Category Discovery}

Vaze et al.~\cite{vaze2022gcd} first learn representations through supervised and self-supervised contrastive objectives, then apply semi-supervised $k$-means for category discovery. SimGCD~\cite{wen2023simgcd} unifies representation learning and parametric classification through self-distillation and entropy regularization. This joint learning framework underlies many recent visual GCD methods~\cite{wang2024sptnet,cao2024legogcd,ma2025protogcd,cao2025allgcd} and also serves as the basis of HeteGCD. These visual methods primarily focus on instance features, leaving topological relations among samples insufficiently explored.

Graph GCD incorporates these relations into category discovery. Deng et al.~\cite{deng2025ggcd} study parametric GCD on graphs and introduce SWIRL for graph contrastive learning. Related work on open-world node classification includes POWN~\cite{hoffmann2025pown}, which combines prototype learning with pseudo-label propagation, and TopoNC~\cite{xi2026toponc}, which selects pseudo-labels based on neighborhood consensus. Such homophily-based mechanisms rely on category agreement between connected nodes as evidence for discovery. On heterophilous graphs, frequent cross-class edges can make this evidence unreliable, motivating other ways to use graph relations.

\subsection{Learning on Heterophilous Graphs}

To move beyond standard GCN aggregation~\cite{kipf2017gcn}, heterophilous graph learning draws on a wider range of attribute and structural signals. Existing methods preserve node attributes and exploit multiple hops~\cite{chen2020gcnii,zhu2020h2gcn} or select and transform hop-wise features through adaptive fusion~\cite{maurya2022fsgnn,gu2025ignn,ghazizadeh2025gamma}. Adaptive filtering and channel mixing capture local differences and diverse frequency responses~\cite{bo2021fagcn,chien2021gprgnn,he2021bernnet,luan2022acm}. Other approaches distinguish homophilous from heterophilous interactions~\cite{du2022gbkgnn} or encode geometric neighborhoods, global relations, and explicit adjacency~\cite{pei2020geomgcn,li2022glognn,lim2021linkx}.

Neighborhood class distributions provide another relational signal: nodes of the same class may have similar neighborhood compositions despite frequent cross-class edges. ND-GCN~\cite{liu2025ndgcn} and NLDGNN~\cite{zhao2026nldgnn} use estimated distributions to guide propagation or construct representations, while SCNode~\cite{uddin2025scnode} incorporates class histograms into node coordinates.

Using these signals effectively requires appropriate selection and fusion, yet GCD provides no direct novel-class supervision to guide this process. Existing neighborhood distribution methods mainly use these statistics for message passing or representation learning, while HeteGCD uses them for category assignment in probability space instead.

\section{Problem Formulation}
\label{sec:problem}

\paragraph{Graph notation}
Let $G=(V,E,X)$ be an undirected attributed graph with $n=|V|$ nodes and $X\in\mathbb{R}^{n\times d_x}$, where row $x_i$ gives the attributes of node $i$ with category $y_i$. We use the row-normalized adjacency matrix $A$ without self-loops, with zero rows for isolated nodes.

\paragraph{Generalized category discovery}
The category set $\mathcal{C}$ comprises disjoint known (old) and novel (new) subsets $\mathcal{C}_o$ and $\mathcal{C}_n$, of sizes $K_o$ and $K_n$. The labeled set $\mathcal{L}\subseteq V$ contains examples from every known category and no novel categories. The unlabeled set $\mathcal{U}=V\setminus\mathcal{L}$ contains nodes from both known and novel categories. GCD aims to assign nodes in $\mathcal{U}$ to categories, jointly recognizing known classes and separating novel classes.

\paragraph{Learning setting}
We consider transductive node-level GCD with known $K=K_o+K_n$. All node features and edges are available during training, while training labels are restricted to $\mathcal{L}$. Hyperparameter and checkpoint selection use only held-out known-class validation labels. Data splits and evaluation details are specified in Section~\ref{sec:experimental-setup} and Appendices~\ref{app:datasets} and~\ref{app:experimental-setup}.

\section{Motivation and Observations}
\label{sec:motivation}

To address the key question raised in the Introduction, we examine neighborhood information from a statistical perspective. With node attributes as the basis, we analyze local attribute-difference statistics and neighborhood class distribution statistics for evidence of category separability.

We assess category information using two measurements. Class separation is the ratio of between- to within-class distances, minus a shuffled-label baseline. Leave-one-out (LOO) nearest-prototype balanced accuracy assigns each node to the nearest class mean after excluding itself, with accuracy averaged equally across classes. Heatmaps visualize within- and between-class similarity. Details are provided in Appendices~\ref{app:raw-observation-metrics} and~\ref{app:context-observation-metrics}.

\subsection{Node Attributes}
\label{sec:raw-attributes}

\paragraph{Intuition}
Since heterophily concerns graph connectivity, raw node attributes can still carry category information before neighborhood aggregation.

\paragraph{Analysis}
Retaining $X$ preserves direct access to each node's own attributes. H2GCN, LINKX, and ACM-GCN likewise treat raw attributes as a key source of category information~\cite{zhu2020h2gcn,lim2021linkx,luan2022acm}.
\begin{figure}[t]
  \centering
  \begin{subfigure}[t]{0.58\columnwidth}
    \centering
    \includegraphics[width=.92\linewidth,trim=0bp 10bp 0bp 8.5bp,clip]{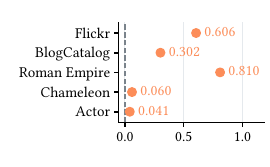}
    \captionsetup{justification=centering,singlelinecheck=false}
    \caption{Class separation above\\the shuffled baseline.}
    \label{fig:raw-x-geometry}
  \end{subfigure}\hfill
  \begin{subfigure}[t]{0.40\columnwidth}
    \centering
    \includegraphics[width=.92\linewidth,trim=0bp 10bp 0bp 8.5bp,clip]{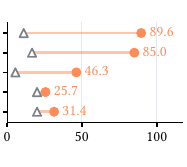}
    \captionsetup{justification=centering,singlelinecheck=false}
    \caption{Nearest-prototype\\classification accuracy (\%).}
    \label{fig:raw-x-loo}
  \end{subfigure}

\par
\begin{subfigure}[t]{.326\columnwidth}\centering
\includegraphics[width=\linewidth,trim=0bp 7bp 0bp 8.5bp,clip]{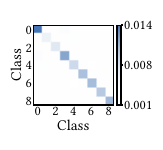}
\caption{Flickr.}\label{fig:raw-x-flickr}\end{subfigure}%
\hfill%
\begin{subfigure}[t]{.326\columnwidth}\centering
\includegraphics[width=\linewidth,trim=0bp 7bp 0bp 8.5bp,clip]{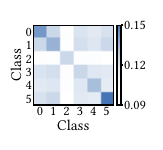}
\caption{BlogCatalog.}\label{fig:raw-x-blogcatalog}\end{subfigure}%
\hfill%
\begin{subfigure}[t]{.326\columnwidth}\centering
\includegraphics[width=\linewidth,trim=0bp 7bp 0bp 8.5bp,clip]{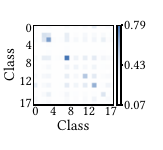}
\caption{Roman Empire.}\label{fig:raw-x-roman}\end{subfigure}
  \caption{Raw node attributes $X$ retain category information across all five datasets. (a) Class separation above the shuffled-label baseline. (b) Leave-one-out nearest-prototype balanced accuracy. Triangles mark chance levels. (c--e) Mean cosine similarity between node pairs, grouped by class.}
  \Description{Five panels show class geometry, leave-one-out assignment, and three heatmaps ordered Flickr, BlogCatalog, Roman Empire from left to right. Every heatmap has an independent vertical colorbar on the right and excludes node self-pairs.}
  \label{fig:raw-x}
\end{figure}

\paragraph{Empirical observations}
Figures~\ref{fig:raw-x-geometry} and~\ref{fig:raw-x-loo} show positive geometry excess and above-chance LOO balanced accuracy across all five datasets. The heatmaps in Figures~\ref{fig:raw-x-flickr}--\ref{fig:raw-x-roman} also show higher average similarity within categories than between categories. Together, these results support retaining $X$ as a direct source of category information for discovery.

\subsection{Local Attribute-Difference Statistics}
\label{sec:local-attribute-differences}

\paragraph{Intuition}
On heterophilous graphs, neighbors often belong to different categories. A node's average attribute differences from its neighbors may nevertheless exhibit category-related patterns.

\paragraph{Analysis}
With the adjacency convention in Section~\ref{sec:problem}, the local attribute-difference signal is $R=X-AX$. For a node with a nonempty neighborhood $\mathcal{N}(i)$, we compute
\begin{equation}
  r_i=x_i-\frac{1}{|\mathcal{N}(i)|}\sum_{j\in\mathcal{N}(i)}x_j
     =\frac{1}{|\mathcal{N}(i)|}\sum_{j\in\mathcal{N}(i)}(x_i-x_j).
  \label{eq:attribute-residual}
\end{equation}
This first-order statistic averages the signed differences $x_i-x_j$ over neighbors, retaining attribute coordinates without category labels. The usefulness of local differences for graph representation learning is also supported by FAGCN, ACM-GCN, and GREET~\cite{bo2021fagcn,luan2022acm,liu2023greet}.

\begin{figure}[t]
  \centering
  \begin{subfigure}[t]{0.58\columnwidth}
    \centering
    \includegraphics[width=.92\linewidth,trim=0bp 10bp 0bp 8.5bp,clip]{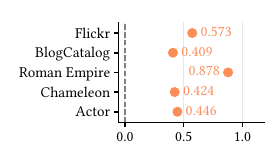}
    \captionsetup{justification=centering,singlelinecheck=false}
    \caption{Class separation above\\the shuffled baseline.}    
    \label{fig:residual-geometry}
  \end{subfigure}\hfill
  \begin{subfigure}[t]{0.40\columnwidth}
    \centering
    \includegraphics[width=.92\linewidth,trim=0bp 10bp 0bp 8.5bp,clip]{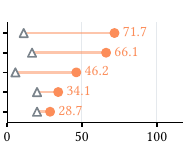}
    \captionsetup{justification=centering,singlelinecheck=false}
    \caption{Nearest-prototype\\classification accuracy (\%).}
    \label{fig:residual-loo}
  \end{subfigure}

\par
\begin{subfigure}[t]{.326\columnwidth}\centering
\includegraphics[width=\linewidth,trim=0bp 7bp 0bp 8.5bp,clip]{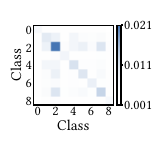}
\caption{Flickr.}\label{fig:residual-flickr}\end{subfigure}%
\hfill%
\begin{subfigure}[t]{.326\columnwidth}\centering
\includegraphics[width=\linewidth,trim=0bp 7bp 0bp 8.5bp,clip]{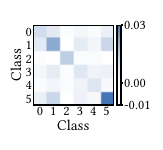}
\caption{BlogCatalog.}\label{fig:residual-blogcatalog}\end{subfigure}%
\hfill%
\begin{subfigure}[t]{.326\columnwidth}\centering
\includegraphics[width=\linewidth,trim=0bp 7bp 0bp 8.5bp,clip]{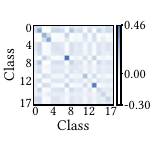}
\caption{Roman Empire.}\label{fig:residual-roman}\end{subfigure}
  \caption{Local attribute-difference statistics $X-AX$ retain category information across all five datasets. (a) Class separation above the shuffled-label baseline. (b) LOO balanced accuracy. Triangles mark chance levels. (c--e) Mean cosine similarity between local difference vectors, grouped by class.}
  \Description{Five panels show class geometry, leave-one-out assignment, and three heatmaps ordered Flickr, BlogCatalog, Roman Empire from left to right. Every heatmap has an independent vertical colorbar on the right and excludes node self-pairs.}
  \label{fig:residual}
\end{figure}

\paragraph{Empirical observations}
This statistic also shows positive geometry excess and above-chance LOO balanced accuracy on all five datasets (Figures~\ref{fig:residual-geometry} and~\ref{fig:residual-loo}). The heatmaps in Figures~\ref{fig:residual-flickr}--\ref{fig:residual-roman} show higher average similarity within categories than between categories. These results support $X-AX$ as an additional source of category information.

\subsection{Neighborhood Class Distribution Statistics}
\label{sec:context-observations}

\paragraph{Intuition}
Even on heterophilous graphs, nodes of the same category can connect to similar mixtures of categories. The relative frequencies of neighbor categories summarize these mixtures and can therefore provide clues to a node's category.

\paragraph{Analysis}
Let $\mathcal{N}_h(i)$ denote the nodes at shortest-path distance exactly $h$ from node $i$. For a nonempty neighborhood, define the empirical neighborhood class distribution and its class mean as
\begin{align}
  q_i^{(h)}(k)
    &=\frac{1}{|\mathcal{N}_h(i)|}
      \sum_{j\in\mathcal{N}_h(i)}\mathbf{1}\{y_j=k\},
      \label{eq:context-composition}\\
  \mu_c^{(h)}
    &=\frac{1}{|V_c|}\sum_{i\in V_c}q_i^{(h)},
      \qquad V_c=\{i:y_i=c\}.
      \label{eq:context-prototype}
\end{align}
Here, $q_i^{(h)}$ records empirical category proportions, and the class mean $\mu_c^{(h)}$ summarizes typical neighborhood composition as a \emph{distribution prototype}. We compare distributions using total variation distance, $d_{\mathrm{TV}}(p,q)=\frac12\lVert p-q\rVert_1$.

Prior work likewise identifies neighborhood category distributions as informative signals for distinguishing node categories~\cite{ma2022homophily,uddin2025scnode,zhao2026nldgnn}, while our focus is their value for discovering novel categories without direct supervision.

\begin{figure}[!t]
\centering
\includegraphics[width=.99\columnwidth,trim=0bp 1.5bp 0bp 1.5bp,clip]{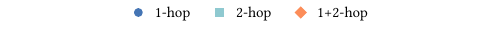}\par
\begin{subfigure}[t]{.58\columnwidth}
  \centering
  \includegraphics[width=.92\linewidth,trim=0bp 10bp 0bp 8.5bp,clip]{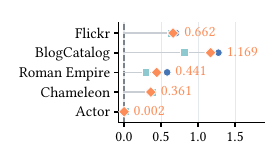}
    \captionsetup{justification=centering,singlelinecheck=false}
    \caption{Class separation above\\the shuffled baseline.}
    \label{fig:context-fusion-geometry}
\end{subfigure}%
\hfill%
\begin{subfigure}[t]{.40\columnwidth}
  \centering
  \includegraphics[width=.92\linewidth,trim=0bp 10bp 0bp 8.5bp,clip]{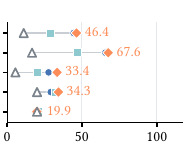}
    \captionsetup{justification=centering,singlelinecheck=false}
    \caption{Nearest-prototype\\classification accuracy (\%).}
    \label{fig:context-fusion-loo}
\end{subfigure}
\par
\begin{subfigure}[t]{.326\columnwidth}
  \centering
  \includegraphics[width=\linewidth,trim=0bp 7bp 0bp 8.5bp,clip]{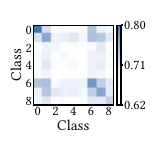}
  \caption{Flickr.}\label{fig:context-fusion-flickr}
\end{subfigure}%
\hfill%
\begin{subfigure}[t]{.326\columnwidth}
  \centering
  \includegraphics[width=\linewidth,trim=0bp 7bp 0bp 8.5bp,clip]{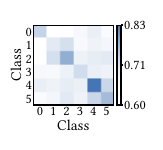}
  \caption{BlogCatalog.}\label{fig:context-fusion-blogcatalog}
\end{subfigure}%
\hfill%
\begin{subfigure}[t]{.326\columnwidth}
  \centering
  \includegraphics[width=\linewidth,trim=0bp 7bp 0bp 8.5bp,clip]{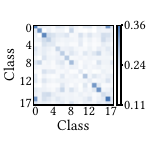}
  \caption{Roman Empire.}\label{fig:context-fusion-roman_empire}
\end{subfigure}
\caption{Combining 1-hop and 2-hop distributions gives higher LOO accuracy than either hop on four of five datasets. (a,b) Class separation and LOO balanced accuracy. Triangles mark chance levels. (c--e) Mean node-pair similarity $1-d_{\mathrm{TV}}$, averaged across hops and grouped by class.}
\label{fig:context-fusion}
\Description{Five panels show geometry, leave-one-out separability, and class-pair heatmaps for fused one- and two-hop neighborhood class distributions. The heatmaps run Flickr, BlogCatalog, Roman Empire from left to right, with independent vertical colorbars.}
\end{figure}

\paragraph{Empirical observations}
Figure~\ref{fig:context-fusion} compares neighborhood class distributions at exact 1-hop and 2-hop distances and their fusion on the same five datasets. To combine the two scales, we retain hop-specific prototypes and average their matching distances:
\begin{equation}
  d_{1+2}(i,c)=\frac12\sum_{h=1}^{2}
                 d_{\mathrm{TV}}(q_i^{(h)},\mu_c^{(h)}).
  \label{eq:context-fused-distance}
\end{equation}
Both hops show positive geometry excess and above-chance LOO balanced accuracy on most datasets, indicating useful category information at both scales. Their relative usefulness varies, 1-hop gives higher LOO accuracy on Flickr and BlogCatalog, whereas 2-hop performs better on Chameleon. Separate analyses of the two hops are provided in Appendix~\ref{app:single-hop-observations}. Fusion exceeds both individual hops on four datasets, and its heatmaps show higher average similarity within categories than between categories. These observations support using both hops to capture complementary neighborhood category information.

In GCD, category assignments for unlabeled nodes must be estimated and may be inaccurate. Such errors alter the estimated neighborhood composition. We therefore examine whether these statistics retain category separability under corrupted assignments.

\begin{figure}[!t]
\centering
\includegraphics[width=.99\columnwidth,trim=0bp 2bp 0bp 1bp,clip]{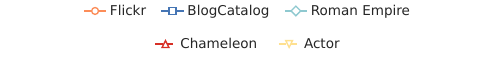}\par
\begin{subfigure}[t]{.495\columnwidth}
  \centering
  \includegraphics[width=\linewidth,trim=0bp 7bp 0bp 1.25bp,clip]{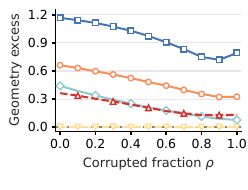}
  \captionsetup{justification=centering,singlelinecheck=false}
  \caption{Class separation above\\the shuffled baseline.}\label{fig:context-random-noise-geometry}
\end{subfigure}%
\hfill%
\begin{subfigure}[t]{.495\columnwidth}
  \centering
  \includegraphics[width=\linewidth,trim=0bp 7bp 0bp 1.25bp,clip]{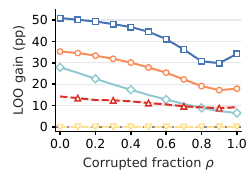}
  \captionsetup{justification=centering,singlelinecheck=false}
  \caption{Nearest-prototype\\accuracy gain over chance (pp).}\label{fig:context-random-noise-loo}
\end{subfigure}
\caption{Fused 1+2-hop distributions retain category information on four of five datasets across the tested noise levels. A fraction $\rho$ of unlabeled nodes per class receives random incorrect labels. (a) Class separation. (b) LOO balanced-accuracy gain over chance (percentage points).}
\label{fig:context-random-noise}
\Description{Random wrong-class noise: geometry excess and LOO gain over chance for Flickr, BlogCatalog, Roman Empire, Chameleon, and Actor. Separate plots share dataset colors and markers; all classes are evaluated with fused one- and two-hop neighborhood class distributions.}
\end{figure}

\paragraph{Random assignment noise} Across the evaluated noise levels, fused distributions retain positive geometry excess and above-chance LOO balanced accuracy on four of the five datasets (Figure~\ref{fig:context-random-noise}), indicating that category information persists despite inaccurate neighbor assignments.

\begin{figure}[!t]
\centering
\includegraphics[width=.99\columnwidth,trim=0bp 2bp 0bp 1bp,clip]{figures/04_observations/shared/dataset_legend.pdf}\par
\begin{subfigure}[t]{.495\columnwidth}
  \centering
  \includegraphics[width=\linewidth,trim=0bp 7bp 0bp 1.25bp,clip]{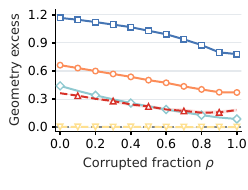}
  \captionsetup{justification=centering,singlelinecheck=false}
  \caption{Class separation above\\the shuffled baseline.}\label{fig:context-old-bias-geometry}
\end{subfigure}%
\hfill%
\begin{subfigure}[t]{.495\columnwidth}
  \centering
  \includegraphics[width=\linewidth,trim=0bp 7bp 0bp 1.25bp,clip]{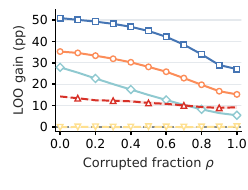}
  \captionsetup{justification=centering,singlelinecheck=false}
  \caption{Nearest-prototype\\accuracy gain over chance (pp).}\label{fig:context-old-bias-loo}
\end{subfigure}
\caption{Four of five datasets retain category information under novel-to-known assignment bias. Corrupted novel-class nodes receive known-class labels. Other settings and measurements follow Figure~\ref{fig:context-random-noise}.}
\label{fig:context-old-bias}
\Description{Novel-to-known assignment bias: geometry excess and LOO gain over chance for Flickr, BlogCatalog, Roman Empire, Chameleon, and Actor. Separate plots share dataset colors and markers; all classes are evaluated with fused one- and two-hop neighborhood class distributions.}
\end{figure}

\paragraph{Novel-to-known assignment bias} We next concentrate erroneous novel-class assignments on known categories, using the same corrupted nodes and the same rule for known-class errors. The same four datasets retain category separability across the evaluated noise levels (Figure~\ref{fig:context-old-bias}). Although separation generally weakens as corruption increases, relational category information remains available under this novel-to-known bias.

These results suggest that neighborhood class distribution statistics can retain useful category information despite assignment noise and thus help guide category assignment in GCD.

\subsection{Design Implications}
\label{sec:observation-summary}

These observations suggest using the signals according to the meaning of their coordinates. $X$ and $X-AX$ share attribute coordinates, describing a node's own properties and its deviations from the neighborhood mean. Their category information lies in patterns of attribute values, which motivates learning an \emph{embedding space} that organizes these complementary cues for category separation.

Neighborhood class distributions already have category semantics, each coordinate records the relative frequency of a particular category in the neighborhood. Existing neighborhood distribution methods primarily use these signals to guide propagation or construct node representations. We instead retain them in \emph{probability space} and match them to distribution prototypes to guide category assignment for both known and novel classes.

\begin{figure*}[t]
  \centering
  \includegraphics[width=.90\textwidth]{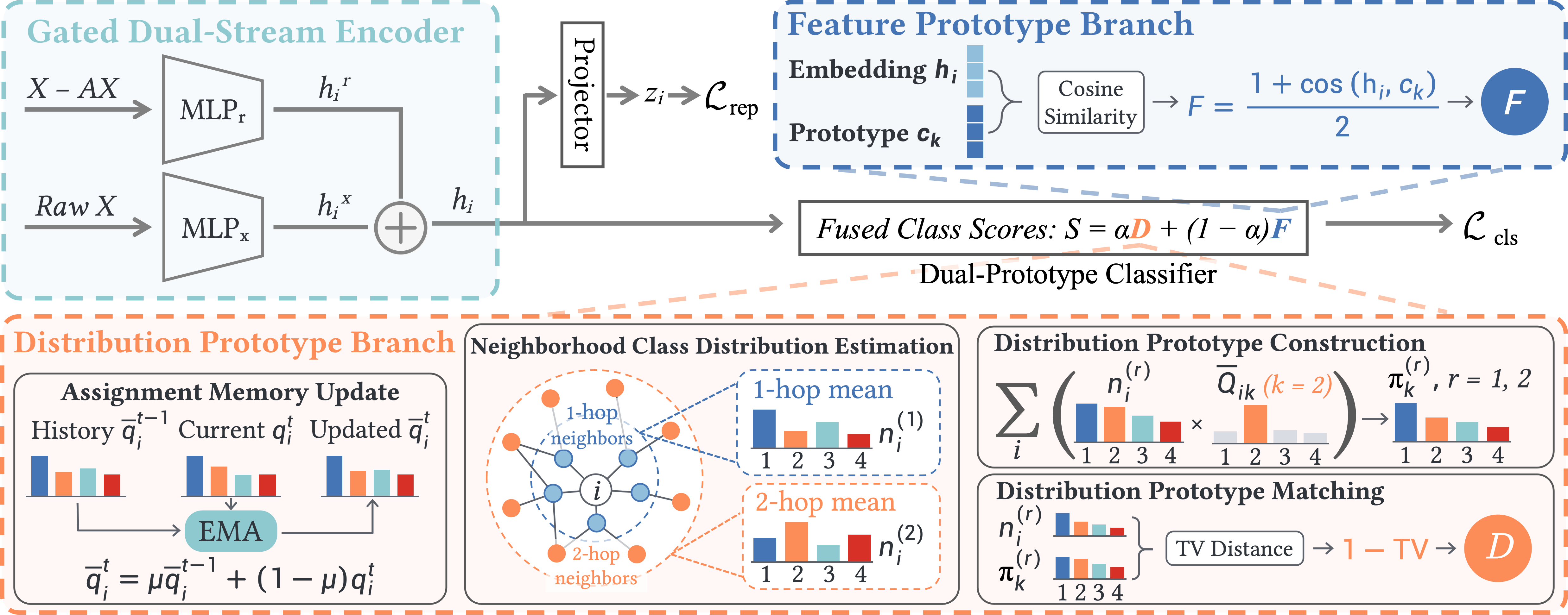}
  \caption{Overview of HeteGCD. A gated dual-stream encoder integrates node attributes $X$ and local attribute-difference statistics $X-AX$ into embeddings $h_i$. Matching $h_i$ to trainable feature prototypes $c_k$ yields scores $F$ in embedding space. Current soft assignments $q_i^t$ update memories $\bar q_i^t$ by exponential moving average. Averaging these memories over exact $r$-hop neighbors estimates neighborhood class distributions $n_i^{(r)}$, $r=1,2$. In probability space, distribution prototype $\pi_k^{(r)}$ is their weighted mean (illustrated for $k=2$), with weight $\bar Q_{ik}$ denoting node $i$'s probability of belonging to class $k$. Total-variation similarities to these prototypes, averaged across hops, yield consistency scores $D$. Fused scores $S$ support classification loss $\mathcal L_{\mathrm{cls}}$, while projected embeddings $z_i$ support contrastive loss $\mathcal L_{\mathrm{rep}}$.}
  \Description{The upper panel shows two independent MLP streams for node attributes and local attribute-difference statistics, their combined embeddings, a contrastive projector, trainable feature prototypes in embedding space, score fusion, and the loss branches. The lower panel updates the soft-assignment memory, estimates neighborhood class distributions over disjoint exact one- and two-hop neighborhoods, and pools these distributions into category-specific distribution prototypes in probability space. Matching node distributions to these prototypes yields total-variation distances, which are converted to detached distribution-based consistency scores.}
  \label{fig:method-framework}
\end{figure*}

Together, these insights motivate a statistics-aware design: combine node attributes and local attribute-difference statistics in embedding space, and match neighborhood class distributions to distribution prototypes in probability space.

\section{Method}
\label{sec:method}

\begingroup
HeteGCD combines a gated dual-stream encoder and a dual-proto\-type classifier to integrate evidence from embedding and probability spaces (Figure~\ref{fig:method-framework}). Appendix~\ref{app:method-pseudocode} provides training and inference pseudocode.
\par
\endgroup

\subsection{Gated Dual-Stream Encoder}
\label{sec:method-encoder}

The encoder combines each node's own attributes with its mean signed attribute differences from neighbors. At each epoch, we generate two graph views by independent feature masking and edge dropping. For view $v$, let $X^{(v)}$ denote the masked attributes and $A^{(v)}$ the row-normalized adjacency without self-loops. We compute local attribute-difference statistics $X_{\mathrm{res}}^{(v)} = X^{(v)} - A^{(v)}X^{(v)}$ and encode the two signals with separate residual MLPs, $f_x$ and $f_r$. A node-wise gate combines their embeddings:
\begin{equation}
\begin{gathered}
  h_i^x=\operatorname{norm}(f_x(x_i)),\quad
  h_i^r=\operatorname{norm}(f_r(r_i)),\\
  g_i=\sigma\!\left(w_g^\top[h_i^x\Vert h_i^r]+b_g\right),\\
  h_i=\operatorname{norm}(h_i^x+g_i h_i^r).
\end{gathered}
\label{eq:method-encoder}
\end{equation}
Here, $x_i$ and $r_i$ are the node's attributes and local attribute-difference statistics in the current view, and $\operatorname{norm}(\cdot)$ denotes $\ell_2$ normalization. The scalar gate $g_i$ controls how much the difference statistics supplement the attribute representation.

A separate projector produces $z_i=\operatorname{norm}(g_{\mathrm{proj}}(h_i))$ for contrastive learning, while classification operates directly on $h_i$.

\subsection{Dual-Prototype Classifier}
\label{sec:method-classifier}

Our classifier combines feature-based similarity with distribution-based consistency for each candidate category. A trainable \emph{feature prototype} represents the category in embedding space, while hop-specific \emph{distribution prototypes} summarize its typical neighborhood class composition in probability space. Feature prototypes are optimized by gradient descent, distribution prototypes are periodically estimated from soft assignments.

\subsubsection{Feature Prototype Matching}
\label{sec:method-feature-prototypes}
Let $c_k$ be a unit-norm feature prototype, as in the cosine classifier of SimGCD~\cite{wen2023simgcd}. We place its matching score on $[0,1]$:
\begin{equation}
  F_{ik}=\frac{1+h_i^\top c_k}{2},\qquad k=1,\ldots,K.
  \label{eq:method-feature-score}
\end{equation}
Before optimization, prototypes are initialized by constrained $k$-means on the initial clean embeddings. Labeled nodes remain in their known slots, and unlabeled nodes can use all $K$ slots. The resulting prototypes are then optimized jointly with the encoder.

\subsubsection{Distribution Prototype Construction and Matching}
\label{sec:method-context-prototypes}
We estimate category-specific neighborhood composition statistics from soft assignments and match each node's distribution to them.

\paragraph{Assignment memory.}
We maintain an assignment memory $\bar{Q} \in \mathbb{R}^{|V| \times K}$, whose row $\bar{q}_i$ stores a smoothed estimate of node $i$'s own class assignment over all $K$ categories. At epoch $t$, for $i \in \mathcal{U}$, the current detached soft assignment $q_i^{(t)}$ updates the memory through an exponential moving average (EMA) with momentum $\mu$:
\begin{equation}
  \bar q_i^{(t)}=\mu\bar q_i^{(t-1)}
     +(1-\mu)q_i^{(t)}.
  \label{eq:method-memory}
\end{equation}
The first update initializes the memory with the current soft assignments, while rows for $i\in\mathcal{L}$ remain fixed to their known-label one-hot vectors.

\paragraph{Neighborhood class distributions.}
On the original graph, let $\mathcal{N}_r(i)$ contain nodes at shortest-path distance exactly $r\in\{1,2\}$ from node $i$. At each refresh, we estimate the neighborhood class proportions by averaging the stored soft assignments:
\begin{equation}
  n_i^{(r)}=\frac{1}{|\mathcal{N}_r(i)|}
             \sum_{j\in\mathcal{N}_r(i)}\bar q_j.
  \label{eq:method-node-context}
\end{equation}
\paragraph{Distribution prototype construction.}
Let $V_r$ index nodes whose $r$-hop neighborhoods are nonempty. For each category $k$, we construct a distribution prototype by averaging these distributions, weighted by the center nodes' soft memberships:
\begin{equation}
  \pi_k^{(r)}=
  \frac{\sum_{i\in V_r}\bar Q_{ik}n_i^{(r)}}
       {\sum_{i\in V_r}\bar Q_{ik}}.
  \label{eq:method-context-prototype}
\end{equation}
Thus $\pi_k^{(r)}$ summarizes the typical neighborhood class composition of nodes assigned to $k$. Both the neighborhood distributions and the center-node weights come from $\bar Q$, allowing this construction to cover known and novel categories without requiring novel-class labels.

\paragraph{Distribution-based consistency.}
We measure consistency between a node's neighborhood composition and that of a candidate category by averaging total-variation (TV) similarities to its distribution prototypes across available hops:
\begin{equation}
  D_{ik}=\frac{1}{|\mathcal{R}_{ik}|}
     \sum_{r\in\mathcal{R}_{ik}}
       \left(1-\frac12\lVert n_i^{(r)}-\pi_k^{(r)}\rVert_1\right),
  \label{eq:method-context-score}
\end{equation}
where $\mathcal{R}_{ik}\subseteq\{1,2\}$ indexes the available hops for this comparison. 

\subsubsection{Score Fusion}
\label{sec:method-score-fusion}
Both $F_{ik}$ and $D_{ik}$ lie in $[0,1]$. We combine each view's feature scores with the shared, detached distribution scores using weight $\alpha$ :
\begin{equation}
  S_{ik}^{(v)}=(1-\alpha)F_{ik}^{(v)}+\alpha D_{ik},
  \qquad v\in\{1,2\}.
  \label{eq:method-score-fusion}
\end{equation}
The fusion weight $\alpha$ is adjusted during training, and the resulting scores $S$ are converted into class probabilities via softmax.

\subsection{Joint Training and Inference}
\label{sec:method-objectives}

We jointly optimize representation and category assignment, using the contrastive objectives of GCD~\cite{vaze2022gcd} and the supervised cross-entropy (CE), self-distillation, and mean-entropy maximization (MeMax) objectives of SimGCD~\cite{wen2023simgcd}. The classification terms use the fused scores from both prototype branches.

\paragraph{Representation learning.}
InfoNCE pairs each node with its other view, and supervised contrastive learning uses labeled nodes of the same category as positives. Both objectives act on projected features $z$:
\begin{equation}
  \mathcal{L}_{\mathrm{rep}}=
  (1-\lambda)\mathcal{L}_{\mathrm{NCE}}
  +\lambda\mathcal{L}_{\mathrm{SupCon}}.
  \label{eq:method-representation-loss}
\end{equation}

\paragraph{Category assignment.}
From the fused scores, we compute student probabilities and sharpened soft targets as
\begin{equation}
\begin{aligned}
  p_i^{(v)}&=\operatorname{softmax}(S_i^{(v)}/\tau_{\mathrm{s}}),\\
  q_i^{(v)}&=\operatorname{sg}\!\left[
                  \operatorname{softmax}(S_i^{(v)}/\tau_{\mathrm{t}})\right],
\end{aligned}
\label{eq:method-probabilities}
\end{equation}
where $\operatorname{sg}$ stops gradients and $\tau_{\mathrm{t}}<\tau_{\mathrm{s}}$. For $\mathcal{L}_k=\{i\in\mathcal{L}:y_i=k\}$, supervised CE is averaged within each known class and then across classes:
\begin{equation}
  \mathcal{L}_{\mathrm{sup}}=-\frac{1}{2K_o}
  \sum_{k=1}^{K_o}\frac{1}{|\mathcal{L}_k|}
  \sum_{i\in\mathcal{L}_k}\sum_{v=1}^{2}\log p_{ik}^{(v)}.
  \label{eq:method-supervised-loss}
\end{equation}
For unlabeled nodes, self-distillation uses each view's soft targets to supervise the other view. With $\ell(q,p)=-q^\top\log p$, the symmetric loss is
\begin{equation}
  \mathcal{L}_{\mathrm{self}}=\frac{1}{2|\mathcal{U}|}
  \sum_{i\in\mathcal{U}}\!\left[
    \ell(q_i^{(2)},p_i^{(1)})+\ell(q_i^{(1)},p_i^{(2)})\right].
  \label{eq:method-self-distillation-loss}
\end{equation}
The MeMax regularizer encourages broad category usage by maximizing the entropy of the mean prediction:
\begin{equation}
\begin{gathered}
  \bar p=\frac{1}{2|\mathcal{U}|}
             \sum_{i\in\mathcal{U}}(p_i^{(1)}+p_i^{(2)}),\\
  H(p)=-\sum_k p_k\log p_k,\\
  \mathcal{L}_{\mathrm{MeMax}}=\log K-H(\bar p).
\end{gathered}
\label{eq:method-memax}
\end{equation}
Using the same supervised weight $\lambda$ in both representation learning and category assignment, the complete objective is
\begin{equation}
\begin{gathered}
  \mathcal{L}_{\mathrm{cls}}=\lambda\mathcal{L}_{\mathrm{sup}}
     +(1-\lambda)\bigl(\mathcal{L}_{\mathrm{self}}
                       +\varepsilon\mathcal{L}_{\mathrm{MeMax}}\bigr),\\
  \mathcal{L}_{\mathrm{total}}=
      \mathcal{L}_{\mathrm{rep}}+\mathcal{L}_{\mathrm{cls}}.
\end{gathered}
\label{eq:method-total-loss}
\end{equation}
\paragraph{Training schedule.}
During warm-up, $\alpha(t)=0$ and predictions use feature prototypes alone, after which $\alpha(t)$ increases linearly to $\alpha_{\max}$. Each epoch predicts and optimizes with the active $D^{(t)}$, then updates $\bar Q$ from the soft targets saved during that forward pass. Distribution scores are first constructed at the end of warm-up and refreshed every five epochs thereafter. For the next epoch, $D^{(t+1)}$ is set to the refreshed scores when available and to $D^{(t)}$ otherwise. At inference, we encode the unaugmented graph and predict $\hat y_i=\arg\max_k S_{ik}$.

\section{Experiments}
\label{sec:experiments}

\subsection{Experimental Setup}
\label{sec:experimental-setup}

\begin{table*}[t]
\centering
\caption{Mean test accuracy (\%) of 25 methods over five settings. Rk. ranks methods by All accuracy within each dataset. Avg. Rank averages these ranks and determines Overall Rank (lower is better). Light orange, light blue, and light cyan indicate the three best distinct values per column, respectively. HeteGCD achieves the best overall rank.}
\label{tab:main-acc-mean}
\begingroup
\ResultTableStyle
\begin{tabular*}{\textwidth}{@{\extracolsep{\fill}}c>{\raggedright\arraybackslash}p{40pt}*{3}{>{\centering\arraybackslash}p{20pt}}>{\centering\arraybackslash}p{13.5pt}*{3}{>{\centering\arraybackslash}p{20pt}}>{\centering\arraybackslash}p{13.5pt}*{3}{>{\centering\arraybackslash}p{20pt}}>{\centering\arraybackslash}p{13.5pt}*{3}{>{\centering\arraybackslash}p{20pt}}>{\centering\arraybackslash}p{13.5pt}*{3}{>{\centering\arraybackslash}p{20pt}}>{\centering\arraybackslash}p{13.5pt}>{\centering\arraybackslash}p{24pt}>{\centering\arraybackslash}p{26pt}@{}}
\toprule
\multicolumn{2}{c}{Method} & \multicolumn{4}{c}{Roman Empire} & \multicolumn{4}{c}{Chameleon} & \multicolumn{4}{c}{Actor} & \multicolumn{4}{c}{BlogCatalog} & \multicolumn{4}{c}{Flickr} & \multicolumn{2}{c}{Summary} \\
\cmidrule(lr){1-2}\cmidrule(lr){3-6}\cmidrule(lr){7-10}\cmidrule(lr){11-14}\cmidrule(lr){15-18}\cmidrule(lr){19-22}\cmidrule(lr){23-24}
 & Encoder & Old & New & All & Rk. & Old & New & All & Rk. & Old & New & All & Rk. & Old & New & All & Rk. & Old & New & All & Rk. & \shortstack{Avg.\\Rank} & \shortstack{Overall\\Rank} \\
\midrule
\multirow{8}{*}{\rotatebox[origin=c]{90}{GCD}} & MLP & 60.89 & 30.92 & 45.58 & 14 & 46.14 & 33.68 & 38.53 & 8 & \MainAccSecond{35.21} & 29.01 & \MainAccSecond{31.78} & \MainAccSecond{2} & 95.16 & \MainAccThird{48.97} & 72.03 & 4 & 92.37 & 13.55 & 48.73 & 11 & \MainAccSecond{7.80} & \MainAccSecond{2.5} \\
 & GCN & 34.50 & 17.98 & 24.52 & 23 & 49.32 & 30.37 & 38.50 & 9 & 20.44 & 30.48 & 27.61 & 22 & 73.03 & 45.07 & 59.12 & 11 & 69.33 & \MainAccFirst{34.86} & 50.29 & 8 & 14.60 & 16 \\
 & GCNII & 71.55 & 27.01 & 48.68 & 6 & 45.80 & 26.66 & 35.91 & 21 & 23.43 & 24.07 & 25.93 & 25 & 97.02 & \MainAccSecond{67.66} & \MainAccSecond{82.32} & \MainAccSecond{2} & 90.91 & 28.54 & 56.36 & 4 & 11.60 & 12 \\
 & H2GCN & 52.87 & 28.81 & 40.61 & 19 & 46.38 & 27.28 & 34.99 & 25 & 33.64 & 23.76 & 28.43 & 20 & 83.32 & 22.08 & 52.66 & 12 & 88.26 & 26.08 & 53.83 & 6 & 16.40 & 18.5 \\
 & LINKX & 58.29 & \MainAccThird{32.90} & 45.03 & 15 & 44.54 & 33.41 & \MainAccSecond{40.06} & \MainAccSecond{2} & \MainAccThird{33.84} & 26.64 & 29.88 & 16 & 97.09 & 25.45 & 61.15 & 9 & 92.41 & 27.58 & \MainAccThird{56.51} & \MainAccThird{3} & 9.00 & 5 \\
 & FSGNN & \MainAccSecond{76.20} & \MainAccFirst{34.40} & \MainAccFirst{54.23} & \MainAccFirst{1} & 42.18 & 31.53 & 37.55 & 13 & 13.20 & 41.94 & \MainAccThird{31.37} & \MainAccThird{3} & 74.65 & 26.30 & 50.65 & 15 & 86.38 & 10.79 & 44.58 & 12 & \MainAccThird{8.80} & \MainAccThird{4} \\
 & ACM-GCN & \MainAccFirst{76.74} & \MainAccSecond{33.13} & \MainAccSecond{54.22} & \MainAccSecond{2} & 47.35 & 30.50 & 37.27 & 14 & 28.45 & 28.64 & 27.97 & 21 & 96.53 & \MainAccFirst{80.91} & \MainAccFirst{88.62} & \MainAccFirst{1} & 92.83 & \MainAccThird{30.55} & \MainAccFirst{58.33} & \MainAccFirst{1} & \MainAccSecond{7.80} & \MainAccSecond{2.5} \\
 & GAMMA & 33.00 & 17.87 & 26.57 & 22 & 36.16 & 34.31 & 36.64 & 18 & \MainAccFirst{38.69} & 21.14 & 29.11 & 18 & 78.29 & 47.79 & 63.28 & 6 & 89.33 & \MainAccSecond{32.88} & \MainAccSecond{58.08} & \MainAccSecond{2} & 13.20 & 15 \\
\midrule
\multirow{8}{*}{\rotatebox[origin=c]{90}{SimGCD}} & MLP & 58.66 & 31.57 & 47.00 & 11 & 45.00 & 33.35 & 38.86 & 6 & 17.53 & 37.51 & 30.68 & 9 & 95.82 & 30.36 & 62.91 & 8 & 94.71 & 3.02 & 43.92 & 14 & 9.60 & 7 \\
 & GCN & 52.57 & 9.30 & 31.02 & 21 & 35.08 & 32.42 & 35.73 & 22 & 13.72 & 33.48 & 26.95 & 24 & 65.91 & 27.77 & 46.93 & 25 & 77.81 & 3.76 & 36.78 & 25 & 23.40 & 25 \\
 & GCNII & 70.01 & 21.94 & 48.53 & 7 & 51.08 & 31.26 & \MainAccThird{39.92} & \MainAccThird{3} & 10.84 & 42.67 & 30.39 & 13 & 79.74 & 20.14 & 50.13 & 18 & 90.47 & 3.64 & 42.35 & 22 & 12.60 & 13 \\
 & H2GCN & 56.06 & 26.50 & 44.13 & 16 & 41.11 & 29.75 & 36.66 & 17 & 17.49 & 36.11 & 29.53 & 17 & 92.05 & 6.84 & 49.31 & 20 & \MainAccSecond{96.09} & 0.00 & 42.86 & 17 & 17.40 & 22 \\
 & LINKX & 63.10 & 31.27 & 47.76 & 9 & 42.27 & 31.78 & 38.32 & 10 & 13.08 & 41.86 & 30.63 & 10 & \MainAccThird{97.84} & 7.31 & 52.34 & 13 & \MainAccThird{95.51} & 11.09 & 48.74 & 10 & 10.40 & 8 \\
 & FSGNN & 70.02 & 21.29 & 46.82 & 13 & 44.19 & 31.80 & 39.29 & 4 & 13.75 & \MainAccFirst{44.30} & \MainAccFirst{32.26} & \MainAccFirst{1} & 84.73 & 13.23 & 49.02 & 21 & \MainAccFirst{96.30} & 0.00 & 42.95 & 15.5 & 10.90 & 9.5 \\
 & ACM-GCN & 65.67 & 28.72 & \MainAccThird{50.60} & \MainAccThird{3} & \MainAccThird{52.46} & 28.20 & 38.85 & 7 & 14.21 & 41.91 & 31.05 & 7 & \MainAccFirst{98.21} & 0.19 & 48.87 & 22 & 91.51 & 3.74 & 42.95 & 15.5 & 10.90 & 9.5 \\
 & GAMMA & 27.24 & 4.28 & 17.78 & 25 & 40.13 & 30.51 & 36.75 & 15 & 8.88 & \MainAccThird{43.42} & 29.91 & 15 & 80.05 & 20.17 & 50.31 & 17 & 91.45 & 6.28 & 44.25 & 13 & 17.00 & 20 \\
\midrule
\multirow{8}{*}{\rotatebox[origin=c]{90}{GGCD}} & MLP & 66.12 & 29.37 & 48.15 & 8 & \MainAccFirst{58.95} & 20.01 & 36.68 & 16 & 14.46 & 41.97 & 31.17 & 5 & 90.28 & 28.80 & 59.44 & 10 & 94.91 & 18.96 & 52.85 & 7 & 9.20 & 6 \\
 & GCN & 56.91 & 7.27 & 32.29 & 20 & 42.37 & 30.98 & 37.86 & 12 & 12.02 & 36.37 & 27.24 & 23 & 70.34 & 30.13 & 50.44 & 16 & 81.73 & 11.14 & 42.66 & 18 & 17.80 & 23 \\
 & GCNII & 66.50 & 23.73 & 47.17 & 10 & 45.30 & 26.84 & 35.49 & 24 & 10.05 & \MainAccSecond{44.13} & 31.17 & 4 & 89.98 & 6.45 & 47.78 & 24 & 93.93 & 0.00 & 41.89 & 24 & 17.20 & 21 \\
 & H2GCN & 56.14 & 25.79 & 42.97 & 17 & 30.27 & \MainAccFirst{36.92} & 36.51 & 19 & 17.73 & 38.04 & 30.49 & 12 & 97.26 & 29.43 & 63.18 & 7 & 90.97 & 3.64 & 42.57 & 19 & 14.80 & 17 \\
 & LINKX & 58.46 & 24.12 & 42.10 & 18 & 34.56 & \MainAccSecond{36.49} & 38.03 & 11 & 26.88 & 33.16 & 30.17 & 14 & 95.11 & 36.38 & 65.54 & 5 & 93.12 & 13.74 & 49.17 & 9 & 11.40 & 11 \\
 & FSGNN & \MainAccThird{72.29} & 19.84 & 46.95 & 12 & 36.78 & 33.31 & 35.57 & 23 & 29.97 & 32.97 & 30.93 & 8 & 82.82 & 15.87 & 49.43 & 19 & 85.74 & 7.64 & 42.52 & 20 & 16.40 & 18.5 \\
 & ACM-GCN & 71.15 & 24.05 & 49.64 & 5 & \MainAccSecond{54.38} & 27.64 & 38.94 & 5 & 14.91 & 40.67 & 30.57 & 11 & \MainAccSecond{98.01} & 0.00 & 48.68 & 23 & 90.45 & 3.60 & 42.40 & 21 & 13.00 & 14 \\
 & GAMMA & 24.56 & 9.27 & 18.13 & 24 & 43.50 & 27.29 & 36.16 & 20 & 22.54 & 28.92 & 28.58 & 19 & 72.35 & 29.37 & 50.90 & 14 & 94.82 & 0.00 & 42.30 & 23 & 20.00 & 24 \\
\midrule
\multicolumn{2}{l}{HeteGCD} & 63.28 & 31.54 & 50.03 & 4 & 47.12 & \MainAccThird{35.94} & \MainAccFirst{40.43} & \MainAccFirst{1} & 14.44 & 41.86 & 31.11 & 6 & 97.29 & 47.85 & \MainAccThird{72.49} & \MainAccThird{3} & 95.32 & 20.50 & 53.90 & 5 & \MainAccFirst{3.80} & \MainAccFirst{1} \\
\bottomrule
\end{tabular*}
\endgroup
\end{table*}

\paragraph{Datasets and protocol.}
We evaluate on five heterophilous graphs, namely Roman Empire~\cite{platonov2023critical}, Chameleon (filtered)~\cite{rozemberczki2021musae,platonov2023critical}, Actor~\cite{pei2020geomgcn}, BlogCatalog~\cite{huang2017lane}, and Flickr~\cite{huang2017lane}. All methods share five frozen class partitions and node splits, with $\lfloor K/2\rfloor$ known classes in each setting. Dataset statistics and split construction are given in Appendix~\ref{app:datasets}.

\paragraph{Comparisons and evaluation.}
We compare against 24 combinations of eight encoders and three discovery frameworks: GCD~\cite{vaze2022gcd}, SimGCD~\cite{wen2023simgcd}, and GGCD~\cite{deng2025ggcd}. The encoders are MLP~\cite{goodfellow2016deep}, GCN~\cite{kipf2017gcn}, GCNII~\cite{chen2020gcnii}, H2GCN~\cite{zhu2020h2gcn}, LINKX~\cite{lim2021linkx}, FSGNN~\cite{maurya2022fsgnn}, ACM-GCN~\cite{luan2022acm}, and GAMMA~\cite{ghazizadeh2025gamma}. Old, New, and All accuracy share a single Hungarian matching over all test nodes.

\paragraph{Implementation.}
All methods use a 100-trial Optuna~\cite{akiba2019optuna} budget and train for 200 epochs. Each trial uses one configuration across the five settings. Hyperparameter and checkpoint selection rely only on known-class validation labels. Appendices~\ref{app:comparison-methods} and~\ref{app:experimental-setup} detail baseline adaptation, search spaces, and the selection rule.

\subsection{Main Results}
\label{sec:overall-performance}

HeteGCD achieves the best average All and New ranks (Table~\ref{tab:main-acc-mean}). Its average All rank is 3.80, compared with 7.80 for both runners-up. It remains in the top six on every dataset, while each baseline ranks 13th or lower on at least one. These results demonstrate strong and consistent overall classification performance across datasets.

New accuracy more directly evaluates novel-class discovery, GCD's central challenge. Here, HeteGCD's average rank is 5.40 versus 8.60 for the runner-up. It ranks in the top eight on every dataset, while each baseline ranks 14th or lower on at least one. These results further demonstrate HeteGCD's ability to discover novel classes across datasets.

\begin{figure}[t]
  \centering
  \includegraphics[width=.80\linewidth,trim=0bp 8bp 0bp 1bp,clip]{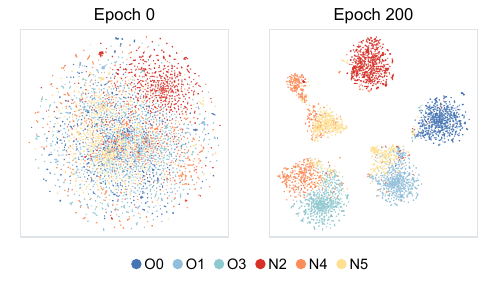}
  \caption{t-SNE visualization of node embeddings before and after training on BlogCatalog. Cool colors denote known-class nodes, and warm colors denote novel-class nodes.}
  \Description{Two t-SNE panels show all 5,196 nodes at epochs 0 and 200. The trained representation exhibits more concentrated category groups and clearer separation.}
  \label{fig:experiment-tsne}
\end{figure}

This consistency is particularly valuable on heterophilous graphs. For example, ACM-GCN+GCD ranks first on BlogCatalog and Flickr but 21st on Actor in All. Strong performance on a few datasets therefore does not necessarily reflect a method's overall effectiveness. Within the SimGCD family, HeteGCD improves All accuracy over MLP+SimGCD on all five datasets, supporting neighborhood statistics as useful complements to node attributes.

\subsection{Embedding Space Visualization}
\label{sec:representation-visualization}

Figure~\ref{fig:experiment-tsne} shows tighter within-category groups and clearer separation between categories in the t-SNE projection~\cite{vandermaaten2008tsne}. This organization is visible for both known and novel categories, illustrating the discriminative embedding space learned by the encoder.

\subsection{Quality of Distribution Prototypes}
\label{sec:context-prototype-quality}

\begin{figure}[t]
  \centering
  \includegraphics[width=\linewidth,trim=0bp 4bp 0bp 13bp,clip]{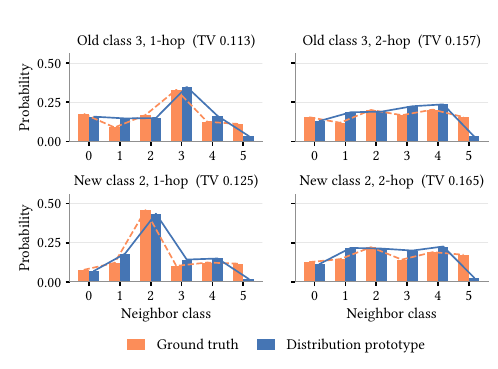}
  \caption{Neighborhood class distributions for a known and a novel class on BlogCatalog at 1-hop and 2-hop. Orange shows the displayed class's ground-truth distribution, while blue shows its estimated distribution prototype.}
  \Description{Four probability-distribution panels compare ground truth and distribution prototypes for known class 3 and novel class 2. The 1-hop and 2-hop TV distances are 0.113 and 0.157 for the known class, and 0.125 and 0.165 for the novel class.}
  \label{fig:context-prototype-case}
\end{figure}

Figure~\ref{fig:context-prototype-case} shows that the estimated distribution prototypes closely follow the main peaks of the ground-truth distributions at 1-hop and capture their broader shapes at 2-hop. This agreement for both known and novel categories indicates that the prototypes encode meaningful category-level neighborhood composition, providing distribution-based evidence for category assignment.

\subsection{Component Ablation}
\label{sec:component-ablation}

Table~\ref{tab:component-ablation} compares Full reruns with variants removing each component under fixed configurations.

\begin{table}[t]
\centering
\caption{Component ablation results with mean test accuracy (\%) over five settings. Avg. Imp. averages Full-minus-ablation differences (pp). Bold marks column maxima.}
\label{tab:component-ablation}
\begingroup\ResultTableStyle
\begin{tabular*}{\linewidth}{@{\extracolsep{\fill}}l*{9}{r}@{}}
\toprule
\multirow{2}{*}{Variant} & \multicolumn{3}{c}{Flickr} & \multicolumn{3}{c}{BlogCatalog} & \multicolumn{3}{c}{Roman Empire} \\
\cmidrule(lr){2-4}\cmidrule(lr){5-7}\cmidrule(lr){8-10}
 & Old & New & All & Old & New & All & Old & New & All \\
\midrule
w/o $X$ & 83.50 & 14.30 & 45.17 & 80.87 & 32.18 & 56.40 & 62.57 & 30.45 & 48.06 \\
w/o $X-AX$ & 95.18 & \textbf{23.91} & \textbf{55.71} & \textbf{97.60} & 49.40 & \textbf{73.41} & 56.45 & 30.07 & 44.79 \\
w/o distribution & 95.00 & 21.10 & 54.09 & 95.56 & \textbf{51.03} & 73.26 & \textbf{70.82} & \textbf{31.90} & \textbf{52.47} \\
Ours (Full) & \textbf{95.41} & 22.89 & 55.28 & 97.56 & 49.13 & 73.28 & 63.52 & 31.14 & 49.99 \\
\midrule
\textit{Avg. Imp.} & +4.19 & +3.12 & +3.62 & +6.22 & +4.93 & +5.59 & +0.24 & +0.34 & +1.55 \\
\bottomrule
\end{tabular*}
\endgroup
\end{table}

Preserving $X$ improves mean All accuracy on all three datasets by up to 16.88 points. $X-AX$ raises Roman Empire's All by 5.19 points, while distribution prototypes raise Flickr's New and BlogCatalog's Old by 1.79 and 2.00 points, respectively. The varying gains are consistent with heterophilous graphs' diverse attribute and connection patterns. Combining attribute information in embedding space with neighborhood compositions in probability space provides complementary evidence for category discovery without novel-class labels. These gains and the best average All and New ranks support HeteGCD's unified design across datasets.

\subsection{Sensitivity to Loss Weights}
\label{sec:loss-sensitivity}

We vary the supervised weight $\lambda$ and the MeMax weight $\varepsilon$ in Eq.~\eqref{eq:method-total-loss} separately on Flickr, BlogCatalog, and Roman Empire.

\begin{figure}[t]
  \centering
  \includegraphics[width=.99\columnwidth]{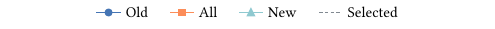}\par
  \begin{subfigure}[t]{0.325\linewidth}
    \centering
    \includegraphics[width=\linewidth]{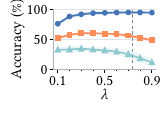}
    \caption{Flickr}
  \end{subfigure}\hfill
  \begin{subfigure}[t]{0.325\linewidth}
    \centering
    \includegraphics[width=\linewidth]{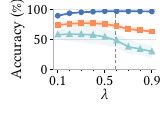}
    \caption{BlogCatalog}
  \end{subfigure}\hfill
  \begin{subfigure}[t]{0.325\linewidth}
    \centering
    \includegraphics[width=\linewidth]{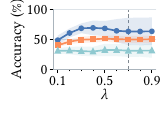}
    \caption{Roman Empire}
  \end{subfigure}

  \caption{Sensitivity to the supervised weight $\lambda$. Lines show mean test accuracy over five settings, and shaded bands indicate $\pm 1$ standard deviation. Dashed vertical lines mark the selected values.}
  \Description{Three separate line plots show Old, New, and All test accuracy against lambda on Flickr, BlogCatalog, and Roman Empire. Old accuracy initially improves. At larger lambda, New accuracy declines on Flickr and BlogCatalog, while it varies less on Roman Empire.}
  \label{fig:sensitivity-lambda}
\end{figure}

\paragraph{Supervised weight.}
Figure~\ref{fig:sensitivity-lambda} shows that increasing $\lambda$ from a low to a moderate value improves Old and All accuracy. At larger values, Old accuracy changes little on Flickr and BlogCatalog, while New accuracy declines. New accuracy on Roman Empire varies less across the tested range.

\begin{figure}[t]
  \centering
  \includegraphics[width=.99\columnwidth]{figures/06_experiments/shared/sensitivity_legend.pdf}\par
  \begin{subfigure}[t]{0.325\linewidth}
    \centering
    \includegraphics[width=\linewidth]{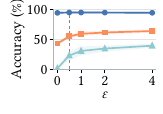}
    \caption{Flickr}
  \end{subfigure}\hfill
  \begin{subfigure}[t]{0.325\linewidth}
    \centering
    \includegraphics[width=\linewidth]{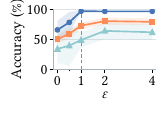}
    \caption{BlogCatalog}
  \end{subfigure}\hfill
  \begin{subfigure}[t]{0.325\linewidth}
    \centering
    \includegraphics[width=\linewidth]{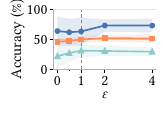}
    \caption{Roman Empire}
  \end{subfigure}

  \caption{Sensitivity to the MeMax weight $\varepsilon$. The plotting conventions follow Figure~\ref{fig:sensitivity-lambda}.}
  \Description{Three separate line plots show Old, New, and All test accuracy against the MeMax weight. Increasing epsilon from zero to two improves mean All accuracy on all three datasets. Further increases have dataset-dependent effects.}
  \label{fig:sensitivity-memax}
\end{figure}

\paragraph{MeMax weight.}
Figure~\ref{fig:sensitivity-memax} shows that increasing $\varepsilon$ from 0 to 2 improves All accuracy on all three datasets, with especially pronounced gains in novel-class discovery on Flickr. Further increases continue to improve New and All accuracy on Flickr, while the gains level off or slightly decline on BlogCatalog and Roman Empire.

Hyperparameters are selected using only known-class validation labels. Consequently, the selected values marked in Figures~\ref{fig:sensitivity-lambda} and~\ref{fig:sensitivity-memax} do not necessarily maximize New or All accuracy.
\section{Conclusion}
\label{sec:conclusion}

In this work, we extend heterophilous graph learning beyond closed-set classification to generalized category discovery. We analyze unreliable homophily priors and missing novel-class supervision, asking how to exploit heterogeneous neighborhood information for discovery. Through empirical observations and analyses from a statistical perspective, we propose HeteGCD, combining a gated dual-stream encoder with a dual-prototype classifier. Experiments on five heterophilous graphs demonstrate its effectiveness. We offer HeteGCD as a pioneering and effective baseline for heterophilous graph GCD and hope to encourage further exploration of heterophilous graph learning in open-world settings.
\bibliographystyle{ACM-Reference-Format}
\bibliography{references}

\appendix

\section{Training and Inference}
\label{app:method-pseudocode}

Algorithm~\ref{alg:simhete-gcd} summarizes HeteGCD. On the original graph, $\mathcal N_r(i)$ is node $i$'s exact $r$-hop neighborhood and $V_r$ contains nodes with nonempty neighborhoods. $\mathcal R_{ik}\subseteq\{1,2\}$ indexes available hops for node $i$ and category slot $k$. Refreshed distribution scores take effect in the next epoch.

\begin{algorithm}[t]
\caption{HeteGCD: training and inference}
\label{alg:simhete-gcd}
\small
\begin{algorithmic}[1]
\Require Graph $G=(V,E,X)$, labeled nodes $(\mathcal L,y_{\mathcal L})$,
  category count $K$, and training epochs $T$.
\Ensure Predictions for unlabeled nodes $\mathcal U$.
\State Initialize encoder $f_\theta$ and projector $g_\phi$;
  initialize feature prototypes $\{c_k\}$ by constrained clustering.
\State Set assignment memory $\bar Q\gets\varnothing$ and distribution scores
  $D^{(1)}\gets0$; use $\alpha(t)=0$ during warm-up.
\For{$t=1,\ldots,T$}
  \Statex \hspace{\algorithmicindent}\textit{Representation and assignment learning}
  \State Generate two views by feature masking and edge dropping.
  \State For each view $v\in\{1,2\}$, compute
    $X_{\mathrm{res}}^{(v)}=X^{(v)}-A^{(v)}X^{(v)}$.
  \State Encode and project both views:
  \Statex \hspace{2em}$H^{(v)}=f_\theta(X^{(v)},X_{\mathrm{res}}^{(v)})$,
    $Z^{(v)}=g_\phi(H^{(v)})$.
  \State Compute feature scores and fuse active distribution scores:
  \Statex \hspace{2em}$F_{ik}^{(v)}=(1+\langle h_i^{(v)},c_k\rangle)/2$,
  \Statex \hspace{2em}$S^{(v)}=(1-\alpha(t))F^{(v)}+\alpha(t) D^{(t)}$.
  \Statex \hspace{2em}Use feature scores alone where distribution scores are unavailable.
  \State Save detached targets
    $Q^{(v)}=\operatorname{sg}[\operatorname{softmax}(S^{(v)}/\tau_{\mathrm{t}}(t))]$.
  \State Take one AdamW step on the joint loss (Section~\ref{sec:method-objectives}).
  \Statex \hspace{\algorithmicindent}\textit{Distribution update}
  \State Initialize or update $\bar Q$ by EMA of $(Q^{(1)}+Q^{(2)})/2$;
    fix labeled rows to their known-label one-hot vectors.
  \State Keep the current distribution scores: $D^{(t+1)}\gets D^{(t)}$.
  \If{warm-up ends or a five-epoch distribution refresh is due}
    \State Estimate neighborhood class distributions for $r\in\{1,2\}$:
    \Statex \hspace{2em}$n_i^{(r)}\gets
      \dfrac{1}{|\mathcal N_r(i)|}
      \sum_{j\in\mathcal N_r(i)}\bar q_j,\quad i\in V_r$.
    \State Construct distribution prototypes for $k=1,\ldots,K$:
    \Statex \hspace{2em}$\pi_k^{(r)}\gets
      \dfrac{\sum_{i\in V_r}\bar Q_{ik}n_i^{(r)}}
            {\sum_{i\in V_r}\bar Q_{ik}}$.
    \State Compute node--slot TV distances:
    \Statex \hspace{2em}$d_{ik}^{(r)}\gets
      \tfrac12\lVert n_i^{(r)}-\pi_k^{(r)}\rVert_1$.
    \State Update distribution scores for the next epoch:
    \Statex \hspace{2em}$(D^{(t+1)})_{ik}\gets
      \dfrac{1}{|\mathcal R_{ik}|}
      \sum_{r\in\mathcal R_{ik}}(1-d_{ik}^{(r)})$.
  \EndIf
\EndFor
\State Select an epoch using Appendix~\ref{app:epoch-selection}.
\State \Return $\hat y_i=\arg\max_k S_{ik}$ for $i\in\mathcal U$,
  using the selected network on the clean graph and its active distribution scores.
\end{algorithmic}
\end{algorithm}

\section{Datasets and Settings}
\label{app:datasets}

\subsection{Datasets and Node Splits}

Table~\ref{tab:dataset-statistics} summarizes the graphs. The split ratios reported below follow the training/validation/test order.

Roman Empire~\cite{platonov2023critical} is a Wikipedia word graph with sequential and syntactic edges, word-embedding features, and syntactic-role labels. We use the first five official 50/25/25 splits.

Chameleon~\cite{rozemberczki2021musae} is a Wikipedia hyperlink graph with page-con\-tent features and five traffic-based categories from Geom-GCN~\cite{pei2020geomgcn}. We use Platonov et al.'s filtered release~\cite{platonov2023critical} and its first five official splits.

Actor~\cite{pei2020geomgcn} is an actor co-occurrence graph from Wikipedia, with keyword features. We use the first five supplied 48/32/20 splits.

BlogCatalog and Flickr~\cite{huang2017lane} are user graphs from blogging and photo-sharing platforms, respectively, with user attributes. We use five fixed, class-stratified 60/20/20 splits per graph.

\begin{table*}[t]
\centering
\caption{Dataset statistics. Labeled ratio gives the minimum--maximum percentage of nodes with training labels across five settings. Edge homophily is the fraction of edges connecting nodes of the same class.}
\label{tab:dataset-statistics}
\small
\begin{tabular*}{\textwidth}{@{\extracolsep{\fill}}ccccccccc@{}}
\toprule
Dataset & Nodes & Edges & Features & Classes & Old Classes & New Classes & labeled ratio (\%) & Edge Homophily \\
\midrule
Roman Empire & 22,662 & 32,927 & 300 & 18 & 9 & 9 & 10.05--29.51 & 0.047 \\
Chameleon & 890 & 8,854 & 2,325 & 5 & 2 & 3 & 16.40--19.55 & 0.236 \\
Actor & 7,600 & 26,659 & 932 & 5 & 2 & 3 & 14.21--23.76 & 0.217 \\
BlogCatalog & 5,196 & 171,743 & 8,189 & 6 & 3 & 3 & 29.08--30.81 & 0.401 \\
Flickr & 7,575 & 239,738 & 12,047 & 9 & 4 & 5 & 26.27--27.29 & 0.239 \\
\bottomrule
\end{tabular*}
\end{table*}

\subsection{Difficulty-Based Settings}
\label{app:data-splits}

We rank candidate known/novel class partitions by cross-group connectivity and feature-centroid proximity between novel and known classes. Stronger connections and closer centroids indicate greater novel-to-known confusability. We select partitions spanning easy, intermediate, and hard regimes and pair them with fixed node splits.

\section{Observation Methodology}
\label{app:observation-metrics}

\subsection{Attribute Signals}
\label{app:raw-observation-metrics}

We examine node attributes $X$ and local attribute-difference statistics $X-AX$, where $A$ averages neighbor attributes without self-loops. For either signal, let $s_i$ be node $i$'s vector and $V_c=\{i:y_i=c\}$ the nodes of class $c$, with $n_c=|V_c|$. The class prototype is $\mu_c=n_c^{-1}\sum_{i\in V_c}s_i$. Comparisons use cosine distance $d(u,v)=1-\cos(u,v)$.

\subsubsection{Class Separation}
$B$ averages distances between distinct class prototypes. $W$ averages node-to-own-prototype distances within each class, then equally across classes:
\begin{equation}
\begin{aligned}
  B&=\frac{2}{K(K-1)}\sum_{c<c'}d(\mu_c,\mu_{c'}),\\
  W&=\frac{1}{K}\sum_{c=1}^{K}\frac{1}{n_c}
       \sum_{i\in V_c}d(s_i,\mu_c),\qquad G=\frac{B}{W}.
\end{aligned}
\label{eq:app-geometry}
\end{equation}
A larger ratio $G$ indicates clearer separation.

For 100 label shuffles preserving class sizes, we recompute prototypes and obtain $G^{(b)}$. The \emph{geometry excess} is
\begin{equation}
  \Delta_{\mathrm{geo}}=G-\frac{1}{100}\sum_{b=1}^{100}G^{(b)}.
  \label{eq:app-geometry-excess}
\end{equation}
Positive values indicate stronger category structure than random grouping.

\subsubsection{Leave-One-Out Balanced Accuracy}
For leave-one-out (LOO) evaluation, we exclude node $i$ when computing class prototypes:
\begin{equation}
  \mu_{y_i}^{(-i)}=
  \frac{n_{y_i}\mu_{y_i}-s_i}{n_{y_i}-1},
  \qquad \mu_c^{(-i)}=\mu_c\quad(c\ne y_i).
  \label{eq:app-loo-prototype}
\end{equation}
Let $a_i$ indicate whether the nearest prototype under cosine distance is correct. Balanced accuracy averages $a_i$ within classes, then equally across classes:
\begin{equation}
  \mathrm{BA}_{\mathrm{LOO}}=
  \frac{1}{K}\sum_{c=1}^{K}\frac{1}{n_c}
  \sum_{i\in V_c}a_i.
  \label{eq:app-loo-ba}
\end{equation}
Plots show this balanced accuracy as a percentage, chance is $1/K$.

\subsection{Neighborhood Class Distributions}
\label{app:context-observation-metrics}

Neighborhood class distributions record the category composition of a node's neighbors. We use exact 1-hop and 2-hop neighborhoods. The 2-hop set excludes the node itself and its immediate neighbors. Write $q_i^{(h)}$ for node $i$'s $h$-hop distribution and $\mu_c^{(h)}$ for the distribution prototype obtained by averaging distributions within true class $c$.

\begin{figure}[!t]
\centering
\begin{subfigure}[t]{.58\columnwidth}
  \centering
  \includegraphics[width=.92\linewidth,trim=0bp 10bp 0bp 8.5bp,clip]{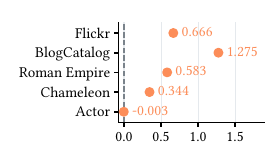}
  \captionsetup{justification=centering,singlelinecheck=false}
  \caption{Class separation above\\the shuffled baseline.}\label{fig:context-1hop-geometry}
\end{subfigure}%
\hfill%
\begin{subfigure}[t]{.40\columnwidth}
  \centering
  \includegraphics[width=.92\linewidth,trim=0bp 10bp 0bp 8.5bp,clip]{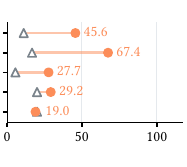}
  \captionsetup{justification=centering,singlelinecheck=false}
  \caption{Nearest-prototype\\classification accuracy (\%).}\label{fig:context-1hop-loo}
\end{subfigure}
\par
\begin{subfigure}[t]{.326\columnwidth}
  \centering
  \includegraphics[width=\linewidth,trim=0bp 7bp 0bp 8.5bp,clip]{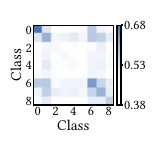}
  \caption{Flickr.}\label{fig:context-1hop-flickr}
\end{subfigure}%
\hfill%
\begin{subfigure}[t]{.326\columnwidth}
  \centering
  \includegraphics[width=\linewidth,trim=0bp 7bp 0bp 8.5bp,clip]{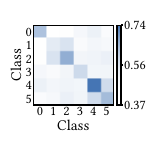}
  \caption{BlogCatalog.}\label{fig:context-1hop-blogcatalog}
\end{subfigure}%
\hfill%
\begin{subfigure}[t]{.326\columnwidth}
  \centering
  \includegraphics[width=\linewidth,trim=0bp 7bp 0bp 8.5bp,clip]{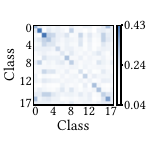}
  \caption{Roman Empire.}\label{fig:context-1hop-roman_empire}
\end{subfigure}
\caption{Exact 1-hop neighborhood class distributions.}
\label{fig:context-1hop}
\Description{Five panels show geometry, leave-one-out separability, and class-pair heatmaps for 1-hop neighborhood class distributions. The heatmaps run Flickr, BlogCatalog, Roman Empire from left to right, with independent vertical colorbars.}
\end{figure}

\subsubsection{Class Separation}
Class separation follows Equation~\eqref{eq:app-geometry}, replacing $s_i$ and $\mu_c$ with $q_i^{(h)}$ and $\mu_c^{(h)}$, and using total variation distance:
\begin{equation}
  d_{\mathrm{TV}}(p,q)=\tfrac12\|p-q\|_1.
  \label{eq:app-tv-distance}
\end{equation}
This gives the between- and within-class distances $B_h$ and $W_h$. For fused distributions, every comparison averages the 1-hop and 2-hop distances, so the geometry ratio is
\begin{equation}
  G_{1+2}=\frac{(B_1+B_2)/2}{(W_1+W_2)/2}
         =\frac{B_1+B_2}{W_1+W_2}.
  \label{eq:app-context-geometry}
\end{equation}
We subtract each signal's shuffled-label reference as in Equation~\eqref{eq:app-geometry-excess}.

For class-pair heatmaps, let $\mathcal P_{ab}=\{(i,j):i\in V_a,\,j\in V_b,\,i\ne j\}$. The mean similarities are
\begin{equation}
\begin{gathered}
  H_{ab}^{(h)}=\frac{1}{|\mathcal P_{ab}|}
    \sum_{(i,j)\in\mathcal P_{ab}}
      [1-d_{\mathrm{TV}}(q_i^{(h)},q_j^{(h)})],\\
  H_{ab}^{(1+2)}=\tfrac12(H_{ab}^{(1)}+H_{ab}^{(2)}).
\end{gathered}
\label{eq:app-context-heatmap}
\end{equation}

\subsubsection{Leave-One-Out Balanced Accuracy}
Following Equation~\eqref{eq:app-loo-prototype}, we exclude each node from its class prototype at each hop to obtain $\mu_c^{(h,-i)}$. The distances are
\begin{equation}
\begin{gathered}
  d_{ic}^{(h)}=d_{\mathrm{TV}}(q_i^{(h)},\mu_c^{(h,-i)}),\\
  d_{ic}^{(1+2)}=\tfrac12(d_{ic}^{(1)}+d_{ic}^{(2)}).
\end{gathered}
\label{eq:app-context-loo-distance}
\end{equation}
Balanced accuracy follows Equation~\eqref{eq:app-loo-ba}.

Clean-observation plots show balanced accuracy as a percentage. Noise plots show the gain over chance in percentage points:
\begin{equation}
  \mathrm{LOO\ gain}=100\bigl(\mathrm{BA}_{\mathrm{LOO}}-1/K\bigr).
  \label{eq:app-loo-gain}
\end{equation}
Zero therefore means chance-level assignment.

\begin{figure}[!t]
\centering
\begin{subfigure}[t]{.58\columnwidth}
  \centering
  \includegraphics[width=.92\linewidth,trim=0bp 10bp 0bp 8.5bp,clip]{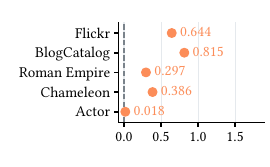}
  \captionsetup{justification=centering,singlelinecheck=false}
  \caption{Class separation above\\the shuffled baseline.}\label{fig:context-2hop-geometry}
\end{subfigure}%
\hfill%
\begin{subfigure}[t]{.40\columnwidth}
  \centering
  \includegraphics[width=.92\linewidth,trim=0bp 10bp 0bp 8.5bp,clip]{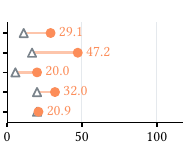}
  \captionsetup{justification=centering,singlelinecheck=false}
  \caption{Nearest-prototype\\classification accuracy (\%).}\label{fig:context-2hop-loo}
\end{subfigure}
\par
\begin{subfigure}[t]{.326\columnwidth}
  \centering
  \includegraphics[width=\linewidth,trim=0bp 7bp 0bp 8.5bp,clip]{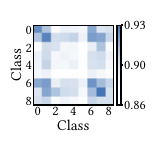}
  \caption{Flickr.}\label{fig:context-2hop-flickr}
\end{subfigure}%
\hfill%
\begin{subfigure}[t]{.326\columnwidth}
  \centering
  \includegraphics[width=\linewidth,trim=0bp 7bp 0bp 8.5bp,clip]{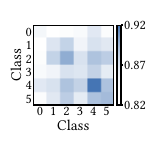}
  \caption{BlogCatalog.}\label{fig:context-2hop-blogcatalog}
\end{subfigure}%
\hfill%
\begin{subfigure}[t]{.326\columnwidth}
  \centering
  \includegraphics[width=\linewidth,trim=0bp 7bp 0bp 8.5bp,clip]{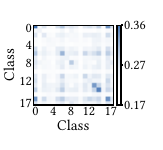}
  \caption{Roman Empire.}\label{fig:context-2hop-roman_empire}
\end{subfigure}
\caption{Exact 2-hop neighborhood class distributions.}
\label{fig:context-2hop}
\Description{Five panels show geometry, leave-one-out separability, and class-pair heatmaps for 2-hop neighborhood class distributions. The heatmaps run Flickr, BlogCatalog, Roman Empire from left to right, with independent vertical colorbars.}
\end{figure}

\subsubsection{Random Assignment Noise}
Random noise tests whether neighborhood class distributions retain category information under incorrect neighbor assignments. For each setting and rate $\rho\in\{0,0.1,\ldots,1\}$, we randomly select a fraction $\rho$ of the unlabeled nodes within each true class. Each selected node receives a uniformly sampled incorrect category.

We rebuild neighborhood distributions from the corrupted assignments, then average them within each true class to obtain class prototypes. LOO accuracy is still evaluated against the nodes' true categories. Each curve averages scores over five settings and 20 repetitions. For the class-separation baseline, we randomly regroup nodes five times per repetition and average the resulting 100 geometry ratios for each setting and noise level.

\subsubsection{Novel-to-Known Assignment Bias}
This bias models novel-class nodes being mistaken for known categories. We corrupt the same nodes as under random noise, assigning selected novel-class nodes uniformly to known categories. Selected known-class nodes receive exactly the same wrong categories as under random noise. We recompute distributions, prototypes, and metrics as above.

\subsection{Separate 1-Hop and 2-Hop Observations}
\label{app:single-hop-observations}

Figures~\ref{fig:context-1hop} and~\ref{fig:context-2hop} provide separate 1-hop and 2-hop results and class-pair heatmaps, complementing the fusion analysis in Figure~\ref{fig:context-fusion}. Neither hop range consistently dominates across datasets, motivating the joint use of neighborhood statistics at different ranges.

\section{Baseline Methods}
\label{app:comparison-methods}

The 24 baselines combine eight encoders with three category discovery frameworks. Encoders extract node representations, while frameworks learn from them and produce category assignments.

\subsection{Graph Encoders}
\label{app:encoders}

MLP~\cite{goodfellow2016deep} transforms each node's attributes independently through feed-forward layers, providing a feature-only reference for assessing the contribution of graph structure.

GCN~\cite{kipf2017gcn} is a canonical message-passing network that provides a reference for conventional neighborhood aggregation. Each layer combines a node's features with normalized neighbor features, smoothing local feature differences.

GCNII~\cite{chen2020gcnii} represents deep graph convolution designed to mitigate oversmoothing. Initial-feature residuals and identity mappings preserve the original node signal during propagation.

H2GCN~\cite{zhu2020h2gcn} is a representative heterophily-oriented encoder that separates ego, 1-hop, and 2-hop information during aggregation. Concatenating representations from different stages retains these sources without immediately mixing neighbors with the node's own features.

LINKX~\cite{lim2021linkx} encodes node attributes and adjacency rows with separate MLPs before combining their outputs, providing a strong non-homophilous graph baseline based on separate attribute and structural encodings.

FSGNN~\cite{maurya2022fsgnn} represents multi-scale feature selection, using precomputed channels from propagation at several depths, with and without self-loops. Separate transformations, learned global weights, and concatenation combine these propagation scales.

ACM-GCN~\cite{luan2022acm} provides a competitive heterophilous graph baseline through adaptive channel mixing. A node-wise mixer combines neighborhood averages, differences from those averages, and ego features, adapting their contributions to each node.

GAMMA~\cite{ghazizadeh2025gamma} is a recent homophily-agnostic graph encoder with adaptive multi-hop aggregation. A shared feature projection propagates over multiple hops, with learned scaling per hop and feature channel. Iterative routing assigns node-specific hop weights based on agreement with the combined representation. We use 0-, 1-, and 2-hop signals with two routing iterations.

\subsection{Category Discovery Frameworks}
\label{app:frameworks}

GCD~\cite{vaze2022gcd} introduced the generalized category discovery setting and provides a foundational clustering-based baseline. Contrastive learning matches two views of the same sample and pulls labeled samples of the same class together. Prediction uses semi-supervised $K$-means on the learned embeddings, keeping labeled assignments fixed to known classes.

SimGCD~\cite{wen2023simgcd} is a widely adopted parametric baseline that jointly learns representations and a prototype classifier through contrastive learning, supervised classification, and cross-view self-distillation, where one view's detached predictions supervise the other. Entropy maximization encourages use of all category slots. Predictions come directly from the prototype classifier.

GGCD~\cite{deng2025ggcd} introduced graph GCD with a focus on homophilous graphs and theoretically analyzed parametric category discovery. Auxiliary semi-supervised clustering and Semantic-aWare dIlation contRastive Learning (SWIRL) regulate category separation using graph structure, alongside contrastive learning, supervised classification, and entropy regularization. The final predictions come from an MLP classifier, while auxiliary clustering supports training.

\subsection{Graph Adaptation and Encoder Integration}
\label{app:baseline-adaptation}

\paragraph{Graph adaptation}
We replace image augmentations with feature masking and undirected edge dropping to construct two graph views~\cite{deng2025ggcd,xi2026toponc}. The same node across views forms a positive pair, supervised contrastive pairs share a known-class label. Edge dropping is disabled for MLP and fixed-support H2GCN.

\paragraph{Encoder integration}
Each encoder is randomly initialized and trained to produce node embeddings $H$, replacing the pretrained image backbone. A projection head maps $H$ to contrastive features. Discovery objectives and prediction remain framework-specific.

\section{Experimental Protocol}
\label{app:experimental-setup}

\subsection{Epoch Selection}
\label{app:epoch-selection}

In the GCD setting, novel-class validation labels are unavailable for model selection or hyperparameter tuning. We therefore use accuracy on known-class validation nodes, comparing predictions directly with known labels without Hungarian matching (Val Old RACC)~\cite{deng2025ggcd}. Known-class validation accuracy may peak while novel-category structure is still developing, motivating later epochs with near-optimal accuracy. Every run trains for 200 epochs. Among observed epochs from epoch 20 onward, we select the \emph{latest} one whose Val Old RACC is at least 97\% of the maximum over these candidates. The same rule is applied to all methods and datasets.

\subsection{Hyperparameter Search}
\label{app:optuna}

Each method--dataset pair has an independent Optuna study~\cite{akiba2019optuna} with 100 trials, including the default configuration. The objective averages validation scores at selected epochs over the five settings. We rerun the best completed trial's configuration on all five settings for final evaluation. Table~\ref{tab:ours-search-space} lists HeteGCD's 14 parameters.

\begin{table}[htbp]
\centering
\caption{Hyperparameter search space for HeteGCD. U and LogU
denote uniform and log-uniform sampling.}
\label{tab:ours-search-space}
\small
\begin{tabular}{lp{.48\linewidth}}
\toprule
Parameter & Search space \\
\midrule
Learning rate & $\mathrm{LogU}[10^{-4},10^{-2}]$ \\
Weight decay & $\{0,10^{-6},10^{-5},3\!\times\!10^{-5},$\newline
               $10^{-4},3\!\times\!10^{-4},10^{-3}\}$ \\
Contrastive batch size & $\{1024,2048,4096\}$ \\
Supervised weight $\lambda$ & $\mathrm{U}[0.1,0.9]$ \\
MeMax weight $\varepsilon$ & $\{0.5,1,2,4\}$ \\
InfoNCE temperature & $\mathrm{LogU}[0.05,1.5]$ \\
SupCon temperature & $\mathrm{LogU}[0.05,0.5]$ \\
Student temperature $\tau_{\mathrm{s}}$ & $\mathrm{LogU}[0.07,0.2]$ \\
Embedding width & $\{64,128,256\}$ \\
Branch dropout & $\{0,0.1,0.2,0.3,0.4,0.5\}$ \\
Feature masking rate & $\{0,0.1,0.2,0.3,0.5\}$ \\
Edge dropping rate & $\{0,0.1,0.2,0.3,0.5\}$ \\
Distribution warm-up fraction & $\{0.1,0.2,0.3\}$ \\
Maximum distribution weight & $\{0.1,0.2,0.3,0.4\}$ \\
\bottomrule
\end{tabular}
\end{table}

\end{document}